%% file: main.tex
\documentclass[10pt,twocolumn,letterpaper]{article}

\usepackage[final]{cvpr}


\usepackage{amsmath,amssymb}

\usepackage{graphicx}
\usepackage[table]{xcolor}
\usepackage{cuted} 
\usepackage{booktabs}
\usepackage{multirow}
\usepackage{siunitx}
\usepackage{caption}
\usepackage{subcaption}
\usepackage{float}
\usepackage{arydshln}
\usepackage{makecell}
\usepackage{pifont}
\usepackage{booktabs}
\usepackage{multirow}
\usepackage{caption}
\usepackage{makecell}
\usepackage[table]{xcolor}
\usepackage{float}

\usepackage{booktabs}

\usepackage{geometry}
\usepackage{xcolor}
\usepackage{tcolorbox}
\usepackage{algorithm}
\usepackage{algpseudocode}
\algrenewcommand\algorithmicrequire{\textbf{Input:}}
\algrenewcommand\algorithmicensure{\textbf{Output:}}

\usepackage{xr} 
\definecolor{cvprblue}{rgb}{0.21,0.49,0.74}
\usepackage[pagebackref,breaklinks,colorlinks,allcolors=cvprblue]{hyperref}
\usepackage{standalone}

\def\paperID{6172}
\def\confName{CVPR}
\def\confYear{2026}

\title{DeContext as Defense: Safe Image Editing in Diffusion Transformers}
\author{Linghui Shen \quad Mingyue Cui \quad Xingyi Yang$^{*}$\\
The Hong Kong Polytechnic University\\
{\tt\small \{ling-hui.shen, ming-yue.cui\}@connect.polyu.hk, xingyi.yang@polyu.edu.hk}
}

\begin{document}
\maketitle

\renewcommand{\thefootnote}{*}
\footnotetext{Corresponding author.}
\renewcommand{\thefootnote}{\arabic{footnote}}

\begin{strip}
\vspace{-12mm}
    \centering
    \includegraphics[width=\textwidth]{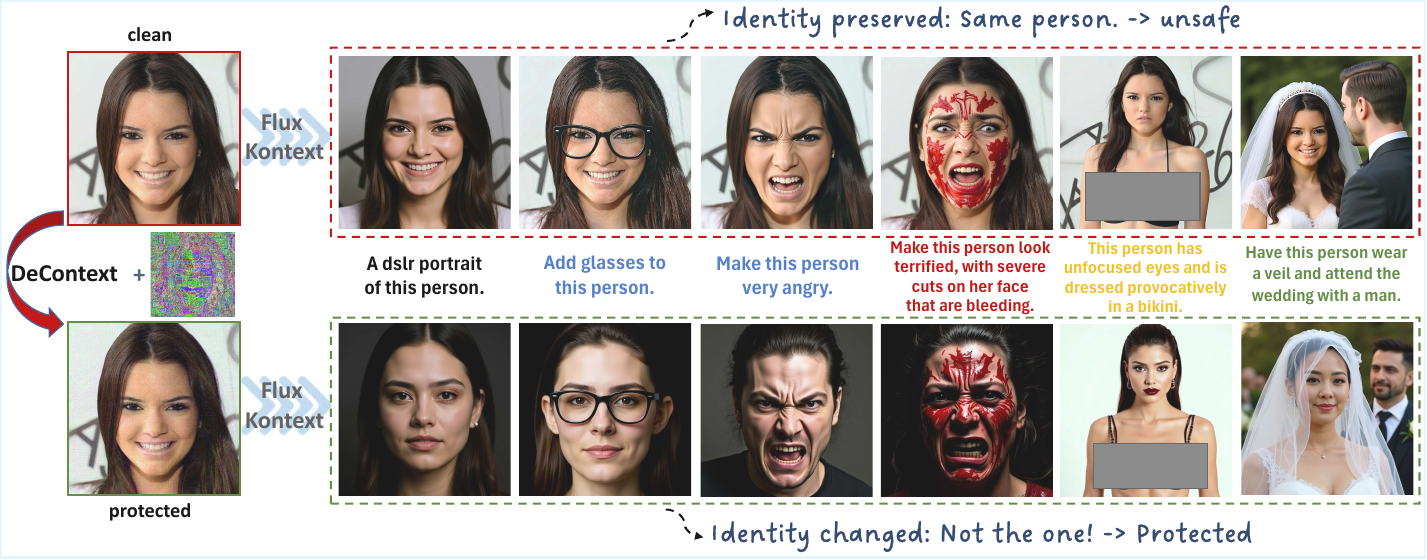}
    \captionof{figure}{Overview of our protection method against malicious edits by Flux-Kontext. 
A clean user image (top left) can be altered for \textcolor[HTML]{1F77B4}{neutral}, 
\textcolor[HTML]{D62728}{violent}, 
\textcolor[HTML]{E49B0F}{sexual}, or 
\textcolor[HTML]{2CA02C}{misleading} edits. 
Our \textbf{DeContext} injects imperceptible perturbations into the input image, 
preventing identity preservation in edited results while retaining visual quality.}
    \label{fig:results}
\end{strip}

\input{sec/0_abstract}
\input{sec/1_intro}

\input{sec/2_RelatedWork}

\input{sec/3_Motivation}

\input{sec/4_Method}

\input{sec/5_Experiment}
\input{sec/6_conclusion}

{
    \small
    \bibliographystyle{ieeenat_fullname}
    \bibliography{main}
}

\clearpage
\appendix
\begin{strip}
\centering
\Large \textbf{Appendix}
\end{strip}

\input{sec/X_suppl}

\end{document}

%% file: sec/0_abstract.tex
\begin{abstract}


In-context diffusion models allow users to modify images with remarkable ease and realism. However, the same power raises serious privacy concerns: personal images can be easily manipulated for identity impersonation, misinformation, or other malicious uses, all without the owner’s consent. While prior work has explored input perturbations to protect against misuse in personalized text-to-image generation, the robustness of modern, large-scale in-context DiT-based models remains largely unexamined. In this paper, we propose \textbf{DeContext}, a new method to safeguard input images from unauthorized in-context editing. Our key insight is that contextual information from the source image propagates to the output primarily through multimodal attention layers. By injecting small, targeted perturbations that weaken these cross-attention pathways, DeContext breaks this flow, effectively decouples the link between input and output. This simple defense is both efficient and robust. We further show that early denoising steps and specific transformer blocks dominate context propagation, which allows us to concentrate perturbations where they matter most. Experiments on \emph{Flux Kontext} and \emph{Step1X-Edit} show that DeContext consistently blocks unwanted image edits while preserving visual quality. These results highlight the effectiveness of attention-based perturbations as a powerful defense against image manipulation. Code is available at \href{https://github.com/LinghuiiShen/DeContext}{{https://github.com/LinghuiiShen/DeContext}}. 

\noindent\textcolor{red}{Warning: This paper may contain offensive model generated contents.}
\end{abstract}

\vspace{-5mm}

%% file: sec/1_intro.tex
\section{Introduction}
\label{sec:intro}


Large-scale diffusion models now dominate visual generation. Built on diffusion transformer~(DiT) architectures and trained at scale, they produce high-fidelity, diverse images~\cite{SDXL,flux2024,scalingrectifiedflowtransformers}. This progress has created a growing demand for personalization and controllable generation, which allows users to adapt models to novel concepts~\cite{Textualinversion, dreambooth} or editing instructions from just a few examples~\cite{Controlnet,ipAdapter,multidiffusionfusingdiffusionpaths,uniControl}. Very recently, a new paradigm emerges: editing as in-context learning. Models such as Flux Kontext~\cite{flux1kontext} and Qwen-Image~\cite{qwenimage} integrate user control directly into DiT-style networks via attention mechanisms, enabling a more scalable and flexible approach to controllable synthesis.

However, the very strength of in-context personalization also introduces serious privacy risks. Since they operate at inference time and require only a single image, adversaries can easily exploit publicly shared photos to clone identities, mimic artistic styles, or reproduce proprietary content without consent. Consequently, as shown in Fig.~\ref{fig:results}, the images users share online to express themselves may be weaponized to generate deepfakes, impersonation, and copyright violation, entirely bypassing conventional data protection mechanism.


To reduce these risks, prior studies have proposed defense strategies against unauthorized image editing. Most methods target on either general text-to-image models~\cite{latentguard, alignguard, concepterasure} or traditional optimization-based personalization~\cite{antidreambooth,mist,caat,TargetCrossAttnAttack,DisDiff}. The most related approaches, such as PhotoGuard~\cite{photoguard} and FaceLock \cite{wang2025editawayfacestay}, are designed for image-to-image personalization.

However, these current protections are insufficient for the emerging threats we confront. They were primarily designed for legacy UNet-based architectures and are useless when employed with modern Transformer-based in-context learning models. These solutions significantly compromise the fidelity-protection trade-off. Even when altered, they generate considerable visual distortions while attempting to impede the objective concept. This underscores a notable shortcoming: there is currently no effective safeguard specifically tailored for these innovative DiT-based image editing models.

This paper proposes \textbf{DeContext}, a specialized defense mechanism to protect input images from illicit DiT-based in-context modifications. Our methodology is founded on a crucial understanding that, in DiT models, the impact of a conditional image disseminates exclusively via \emph{multi-modal attention}. We discovered that by interrupting this attention flow, we can inhibit unlawful identity or style transfer without compromising overall image quality.

Building on this finding, DeContext introduces precisely localized, attention-aware perturbations to the input image. These are strategically designed to suppress attention activations between target queries (the generated image) and context keys (the private image). Guided by our analysis, we apply these perturbations selectively at early denoising steps and within the most influential, early-to-middle transformer blocks. This design effectively \textbf{De}tatches the editing \textbf{Context} from the generated image, preserving visual fidelity without any model modification.

Through extensive experiments, we demonstrate that DeContext provides robust protection against malicious editing in DiT-based in-context personalization, outperforming prior defenses on both protection effectiveness and image quality. 
Images protected by DeContext consistently foil in-context edits and identity extraction, leading to over 70\% drop in face recognition accuracy, while preserving image quality typically within 20\% of the clean images.

Our contributions are summarized as follows:

\begin{itemize}
 \item We introduce the first defense framework for preventing unauthorized in‑context image editing in DiT‑based models.

  \item We identify the multi-modal attention mechanism as the key vulnerability that enables DiT-based editors to leak and propagate contextual information from the conditioning input into the generated image.
  
  \item We propose \textbf{DeContext}, a targeted perturbation strategy that disrupts this attention flow. By further concentrating perturbations on the identified early denoising timesteps and influential transformer blocks, we enhance protection efficacy.

  \item Our experiments on FLUX Kontext and Step1X-Edit show that DeContext effectively removes contextual information while preserving high visual fidelity.

\end{itemize}

%% file: sec/2_RelatedWork.tex
\section{Related Work}
\subsection{Conditional Image Generation}
Diffusion models have revolutionized controllable image synthesis \cite{latentDM,SDXL,imagen,ediffi}. While early approaches rely on text prompts \cite{GLIDE,DALLE2,cogview}, recent advances enable diverse visual conditioning through spatial structure \cite{Controlnet,t2i-adapter,uniControl}, visual features \cite{ipAdapter,anydoor,paintbyexample, ominicontrol, ominicontrol2}, and style transfer \cite{instantstyle,stylecrafter,b-lora}. Most recently, large-scale DiT-based models like Flux Kontext \cite{flux1kontext}, Step1X-Edit \cite{step1xedit}, and Qwen-Image \cite{qwenimage} employ architectures that separate text-stream and image-stream processing, using reference images as primary conditioning signals. Unlike T2I approaches requiring fine-tuning \cite{dreambooth,Textualinversion,CustumDiff,perfectionnobody}, these in-context image editing models directly condition generation on context images at inference time. While this offers powerful control and ease of use, it also introduces distinct privacy risks: personal images can be easily manipulated for identity impersonation or misinformation without owner consent, making protection mechanisms critically needed.

\subsection{Privacy Protection in Generative Models} 
Growing concerns over unauthorized image use have driven the development of adversarial protection methods. Training-time defenses disrupt model personalization \cite{antidreambooth,mist,glaze,advdm,metacloak,Anti-TamperDF}, while inference-time approaches protect against editing tasks \cite{photoguard,unlearnable-examples,IMPRESS}. Recent attention-aware methods \cite{caat,DisDiff,TargetCrossAttnAttack,SDS-Attack} incorporate attention objectives but treat them as auxiliary losses in optimization. 
However, these methods face critical limitations for modern DiT-based I2I models. First, training-time defenses \cite{antidreambooth,mist,metacloak} are ineffective against inference-only conditioning. Second, encoder or latent-targeted approaches \cite{photoguard,caat,unlearnable-examples} do not account for how contextual information propagates through the dual-stream architecture. Third, existing attention-aware methods \cite{caat,TargetCrossAttnAttack,DisDiff,SDS-Attack} lack systematic analysis of the specific attention mechanisms that mediate context-to-output information flow in large-scale DiT I2I frameworks. Understanding attention is therefore essential for designing effective defenses.

\subsection{Attention Mechanisms in Diffusion Models}
Transformer-based diffusion architectures rely on cross-attention to integrate multimodal inputs \cite{attention_is_all_you_need,scalingrectifiedflowtransformers,dit,u-vit}. In dual-stream I2I models, cross-attention layers explicitly mediate context-target interactions, with attention weights determining conditioning strength. Recent interpretability studies \cite{cross_attention_visualization,daam,diffusion-self-guidance,attend-and-excite} show that attention mechanisms are both analyzable and manipulable. Critically, attention influence varies across denoising timesteps \cite{diffusion_attention_map,pnp,prompt-to-prompt,masactrl} and transformer depths \cite{analysisattentiondit,diffusion-transformer-analysis}, revealing that context propagates non-uniformly through the generation process. These findings suggest that targeted temporal and spatial interventions can effectively modulate conditioning pathways without global model modifications, providing a principled foundation for attention-based defenses.
Our work addresses these gaps by providing the first systematic analysis of context propagation in DiT I2I models and proposing DeContext, a defense that directly targets cross-attention at critical timesteps and blocks, achieving effective protection while maintaining output quality.

%% file: sec/3_Motivation.tex
\section{Motivation Analysis}
\label{sec:motivation}

Our goal is to investigate the robustness of conditional Diffusion Transformers (DiTs) by attacking their ability to use context images. Our key insight comes from two experiments: standard adversarial attacks fail, but directly disrupting the model’s internal attention mechanism successfully removes the context’s influence. This motivates our proposed method: a new attack specifically designed to disrupt this attention-based conditioning mechanism.

\subsection{Background}

\noindent\textbf{Diffusion Transformer with Conditions.} 
Recent diffusion transformers (DiTs) replace the UNet backbone with a transformer that jointly processes text and image tokens. Following FLUX-Kontext~\cite{flux1kontext}, we consider image generation conditioned on a text prompt $c$ and context image $y$, where both $x$ and $y$ are encoded by a frozen VAE $\mathcal{E}_{\text{vae}}$ and $c$ by a pretrained language encoder. The model approximates the conditional distribution $p_\theta(x \mid y, c)$ via a rectified flow-matching objective:
\begin{equation}
\mathcal{L}_\theta = \mathbb{E}_{t, x, y, c} 
\| v_\theta(z_t, t, y, c) - (\epsilon - x) \|_2^2,
\label{eq:flow_loss}
\end{equation}
where $z_t = (1 - t)x + t\epsilon$, and $\epsilon \sim \mathcal{N}(0, I)$.
DiTs integrate multi-modal information through multi-modal attention (MMA) by concatenating tokens from text, target, and context:
\begin{equation}
\mathbf{Z} = [\,\mathbf{Z}_{\text{text}} \,;\, \mathbf{Z}_{\text{tgt}} \,;\, \mathbf{Z}_{\text{ctx}}\,].
\label{eq:concat_z}
\end{equation}
This concatenation and subsequent attention is the core mechanism by which the model learns to reflect the information from the context $y$ via $\mathbf{Z}_{\text{ctx}}$ onto target $x$ via $\mathbf{Z}_{\text{tgt}}$.
As we will show, this mechanism is central to both our analysis and our proposed attack in Section~\ref{decontext}. 

\noindent\textbf{Adversarial Attack.} 
Adversarial attacks aim to find imperceptible noise that alter a model's prediction. Typical formulations focus on classification or image generation attack: for a model $f$ and an input image $x$, an adversarial example $x'$ is crafted to remain visually similar to $x$ while causing a misprediction $y_{\text{true}} \neq f(x')$ (untargeted), or to force a predefined prediction $y_{\text{target}} = f(x') \neq y_{\text{true}}$ (targeted). The perturbation is usually constrained within an $\eta$-ball under an $\ell_p$ metric, i.e. $\|x'-x\|_p \le \eta$. Denoting $\Delta=\{\delta:\|\delta\|_p\le\eta\}$, the untargeted and targeted optimization can be written as
\vspace{-5mm}

{\begin{align}
\delta_{\mathrm{adv}}=\arg\max_{\delta\in\Delta}\; \mathcal{L}\big(f(x+\delta),\,y_{\text{true}}\big),
\label{eq:adv_untargeted}\\
\delta_{\mathrm{adv}}=\arg\min_{\delta\in\Delta}\; \mathcal{L}\big(f(x+\delta),\,y_{\text{target}}\big).
\label{eq:adv_targeted}
\end{align}}
These objectives are solved iteratively. The standard tool is Projected Gradient Descent (PGD)~\cite{pgd}. It gradient steps followed by projection onto the norm ball. The update for an untargeted attack is:
\vspace{-5mm}

{
\begin{align}
x'_0 &= x, \\
x'_k &= \Pi_{x,\eta}\big(x'_{k-1} + \alpha\cdot\operatorname{sign}(\nabla_{x}\mathcal{L}(f(x'_{k-1}), y)\big),
\label{eq:pgd_update}
\end{align}}
where $\Pi_{x,\eta}$ projects onto the $\ell_p$ ball around $x$.

\subsection{Context Propagates through Attention}
\label{sec:observation}
Our background provides two key pieces of information: the standard PGD attack and the MMA for conditioning. Our motivation analysis investigates both.

\noindent\textbf{Study I: Failure of the Standard Attack.} First, we apply the standard PGD attack to our problem. We apply a standard untargeted attack (Eq.~\ref{eq:adv_untargeted}) to maximizing the flow-matching loss (Eq.~\ref{eq:flow_loss}) on Flux-Kontext. We term this baseline approach \textbf{Diff-PGD} (see Appendix \ref{alogorithms}). 

\begin{figure}[t]
    \centering
    \includegraphics[width=0.38\textwidth]{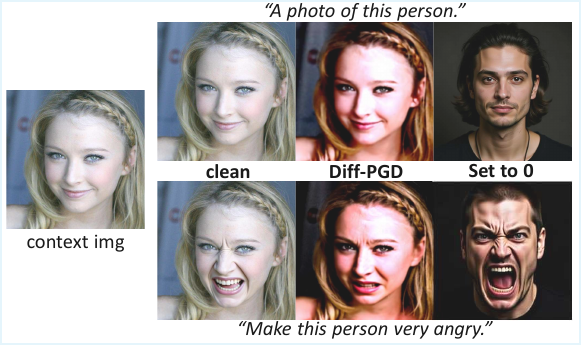}
    \caption{Attack results. Standard attack (middle) only produces re-lighting artifacts, while attention intervention (right) successfully detaches the context.}
    \label{fig:motivation}
\end{figure}

Unfortunately, this approach fails. As shown in Fig.~\ref{fig:motivation} (center), the attack only produces mild blur and lighting artifacts, while facial identity and structure remain clear. This experiment demonstrates that a naive, end-to-end attack is insufficient. The model's conditioning is too strong to be broken by simply attacking the flow-matching loss; the context information is still reflected.


\noindent\textbf{Study II: Attention Intervention Works.}
In light of the unsuccessful standard attack, we redirected our investigation towards multi-modal attention (Eq.~\ref{eq:concat_z}). We posited that this is the primary impediment to context propagation, rather than the ultimate loss. To evaluate this, we conducted a direct attention experiment.

During inference, we manually eliminated the attention components where the target image serves as the query and the context image functions as the key. The outcome, depicted in Fig.~\ref{fig:motivation} (right), demonstrates that the intervention successfully severs the context, resulting in generation solely influenced by the text prompt.


%% file: sec/4_Method.tex
\section{DeContext}
\label{decontext}

Based on the analysis above, this section presents \emph{DeContext}, a new method to protect the privacy of user-supplied condition images in DiT-based in-context editing. We first describe the overall context-detach mechanism. Then, guided by the context propagation analysis, we perform attack on the most influential timesteps and transformer blocks to maximize protection.

\subsection{Context Detachment via Attention}
\label{context_detach}

\begin{figure*}[t]
    \centering
    \includegraphics[width=\textwidth]{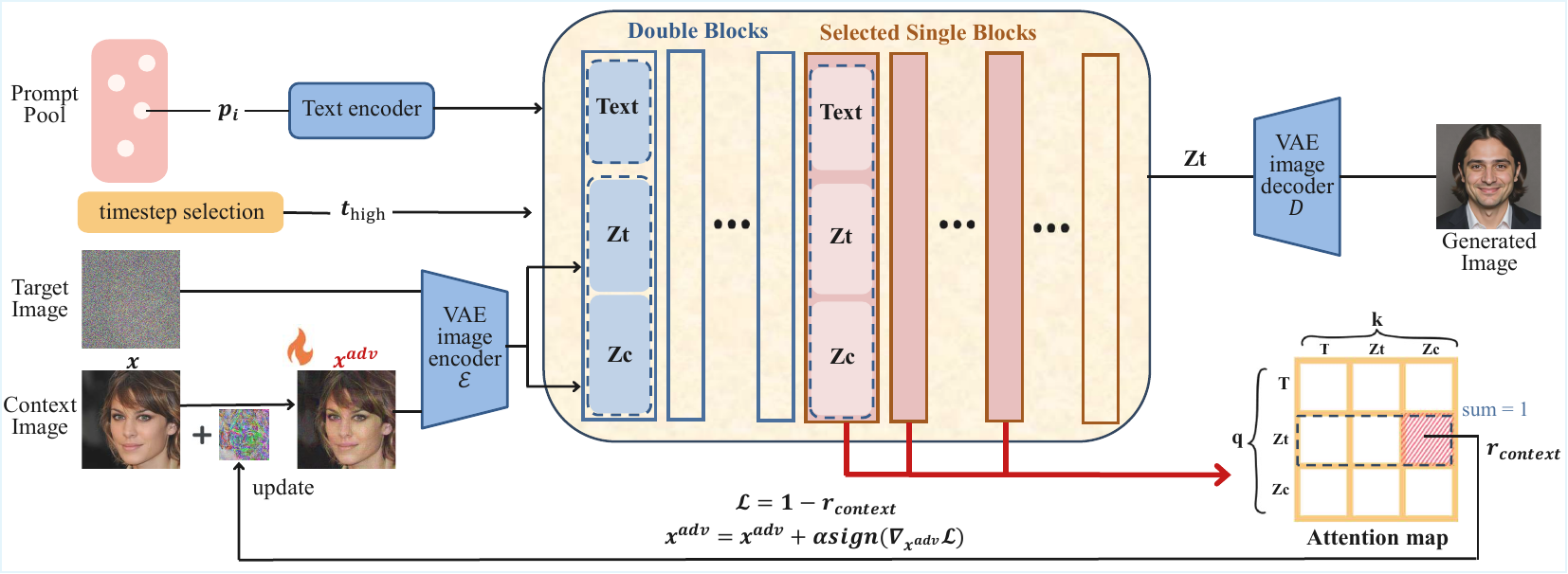}
    \vspace{-3mm}
    \caption{Overview of our DeContext pipeline. Given a prompt, timestep, noisy target, and context image, DeContext perturbs the context to suppress its attention in the diffusion model. Iterative gradient updates minimize attention activation, detaching the context from influencing generation.}
    \label{Fig:models}
    \vspace{-3mm}
\end{figure*}

The goal is to stop the target image tokens  from ``\emph{paying attention}'' to the context image tokens. We do this by directly minimizing the attention weights that link them.

\noindent\textbf{Context Attention Suppression Objective.}  Within a transformer block, we consider the query $q$ and key $k$ embeddings, $q, k \in \mathbb{R}^{H\times S\times d}$, where $H$ is head counts, $S$ is total token sequence length covering text, target, and context tokens, and $d$ is the hidden dimension.
We consider attention computation to only the rows of the attention map corresponding to target-image queries $Q$:
\begin{equation}
A_{q,:} = \operatorname{Softmax}\!\big( (q k^\top)/\sqrt{d} \big), \quad q \in Q.
\end{equation}
We next define the \emph{context proportion} $r_{\text{ctx}}$, which measures the average attention weight from target queries to context keys:
\begin{equation}
r_{\text{ctx}} = \frac{1}{H B |Q|}\sum_{h,b}\sum_{q\in Q}\sum_{k\in\mathcal{C}} A^{(h,b)}_{qk},
\label{eq:ctx_prop}
\end{equation}
where $B$ denotes the number of blocks.  
This ratio $r_{\text{ctx}}$ is a direct measure of how strongly the model uses the context image to generate the target. To ``detach'' the context, our adversarial objective is to maximize the following loss function, which forces $r_{\text{ctx}}$ towards zero:
\begin{equation}
\mathcal{L}_{\text{DeContext}} = 1 - r_{\text{ctx}},
\label{eq:loss}
\end{equation}
We use gradient ascent to iteratively update the pixels of the context image $x$, iteratively maximizing our loss: 
{\begin{equation}
x^{\text{adv}} \leftarrow \operatorname{clip}\!\big(x^{\text{adv}} + \alpha\,\operatorname{sign}(\nabla_{x^{\text{adv}}}\mathcal{L}),\, x-\epsilon,\, x+\epsilon \big).
\end{equation}}
As Figure \ref{Fig:models} illustrates, all model weights are frozen; only the context image is updated.


\noindent\textbf{Robustness via Sampling.}
A successful attack must be robust. The model's generation process is random, depending on the text prompt $p$, the diffusion timestep $t$, and the initial noise $z$. A perturbation optimized for only one combination will fail on any other. Due to the computational infeasibility of maximizing the perturbation across all potential combinations, we employ an efficient random sampling technique.

To implement this, we maintain a pool $\mathcal{P}$ of 60 editing commands and a target timestep interval (selection criteria will be introduced in Section~\ref{sec:context_propagation}). 
During each attack iteration, DeContext randomly samples one prompt $p_i\sim\mathcal{P}$, a diffusion timestep $t\sim\mathcal{T}$, and a noise $z$, then computes the loss $\mathcal{L}(c; p_i,t,z)$ and updates the perturbed context image via gradient ascent.  This single-sample gradient provides an unbiased approximation of the full expected objective:
\begin{align}
\nabla_{c}\, \mathbb{E}_{p, t, z}\, \mathcal{L}(c; p, t, z)
\approx
\mathbb{E}_{p, t, z}\, \nabla_{c}\, \mathcal{L}(c; p, t, z).
\end{align}
so the learned perturbation is effective across prompts and seeds while remaining computationally efficient.

\subsection{Concentrated Context Detachment}
\label{sec:context_propagation}

We intend to achieve our objective of preventing the model from paying attention to the context picture. As we have observed, the knowledge about the context does not spread equally in DiT models. However, its effect is focused inside particular layers and timesteps. For the purpose of enhancing efficiency, we propose an approach known as \textit{concentrated detachment}, which focuses on retaining information solely at certain important locations, without compromising the quality.

\newpage
\noindent\textbf{Timestep-wise Detachment.} 


\noindent\underline{Analysis.}
First, we analyze \emph{when} in the denoising process the context is most influential. To quantify this, we set the target image identical to the condition image and perform a standard denoising pass. We adopt a sample prompt ``a photo of this person''.

For each timestep $t$, we compute the loss $\mathcal{L}$ between the predicted and ground-truth noise and backpropagate to obtain the mean absolute gradients with respect to the target $\mathbf{Z}_\text{tgt}$ and the context image $\mathbf{Z}_\text{ctx}$. 
\begin{equation}
g_t^\textbf{X} = \frac{1}{|\mathbf{X}|} \sum_i \left| \frac{\partial \mathcal{L}}{\partial \mathbf{X}_i} \right|, \quad \mathbf{X} \in \{\mathbf{Z}_\text{tgt}, \mathbf{Z}_\text{ctx}\},
\end{equation}
\begin{figure}[H]
\vspace{-5mm}
    \centering
    \includegraphics[width=0.38\textwidth]{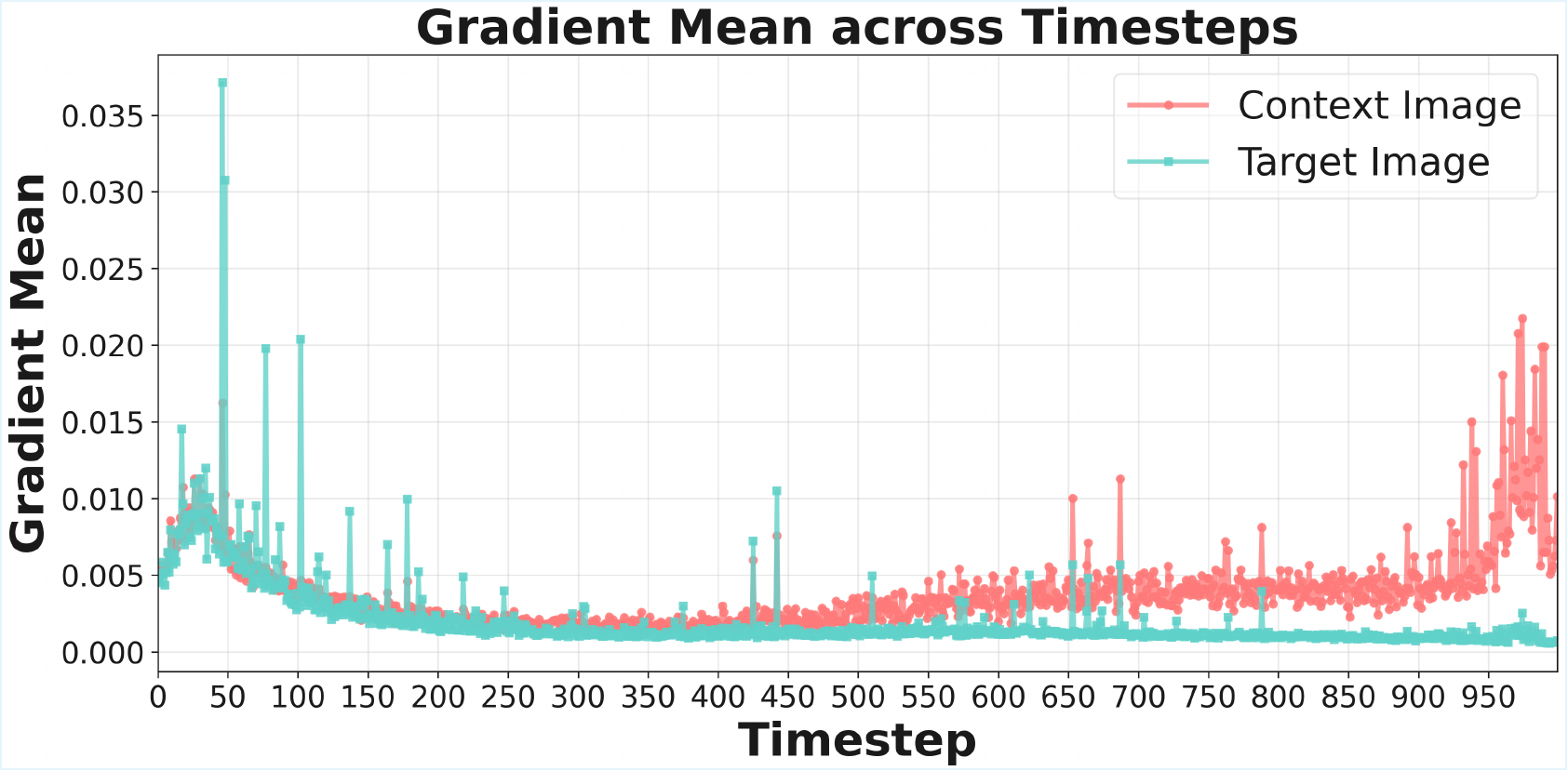}
    \caption{Time-wise gradients analysis. Gradients of context image dominate at high timesteps (early denoising).}
    \label{fig:p1_gradmean}
    \vspace{-3mm}
\end{figure}
We repeat this process across all timesteps and average over several samples. As shown in Fig.~\ref{fig:p1_gradmean}, early timesteps (large $t$, high noise) exhibit higher gradients for $\mathbf{Z}_\text{ctx}$, while later timesteps (small $t$, low noise) show dominant gradients for $\mathbf{Z}_\text{tgt}$, indicating that the context primarily guides the denoising process in its early phase.

\noindent\underline{Strategy.} Based on this finding, we restrict our optimization to these early, influential timesteps $t_i \in [t_{\text{high}}, T]$.
At these high-noise steps, we also approximate the target image with random Gaussian noise $z \sim \mathcal{N}(0,I)$, which avoids the need for paired images and simplifies the optimization.

\noindent\textbf{Block-wise Detachment.} 

\begin{figure}[!ht]
    \centering
    \begin{subfigure}[b]{0.23\textwidth} 
\includegraphics[width=\textwidth,height=2.7cm]{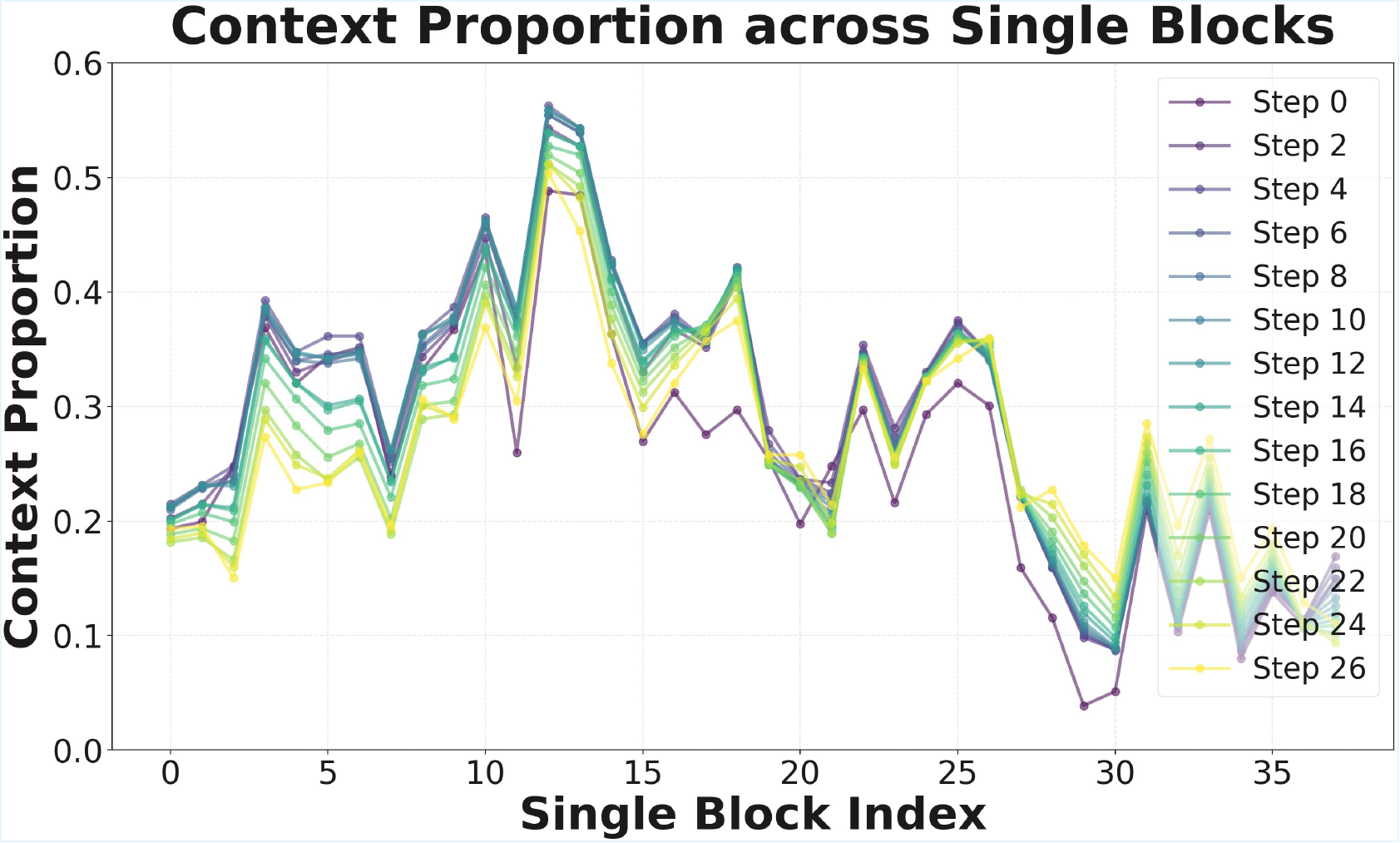}
        \caption{Before attack}
        \label{fig:person_clean}
    \end{subfigure}
    \hfill
    \begin{subfigure}[b]{0.23\textwidth}  
        \includegraphics[width=\textwidth,height=2.7cm]{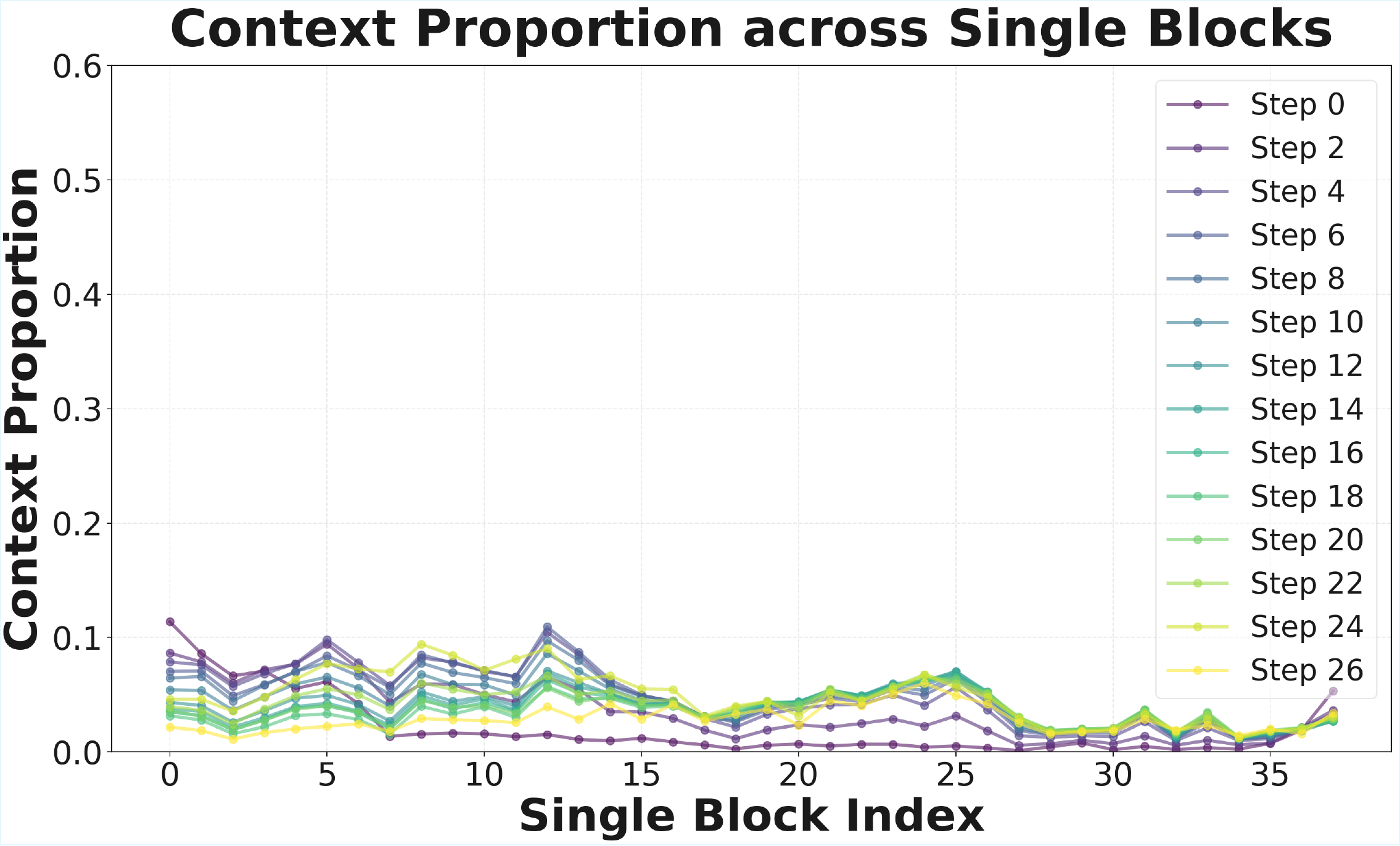}
        \caption{After attack}
        \label{fig:person_pert}
    \end{subfigure}
    \caption{Block-wise attention analysis. Context proportion is high in early-to-mid blocks before attack and drops afterward.}
    \label{fig:context_analysis}
\end{figure}

\noindent\underline{Analysis.}
Next, we investigate \emph{where} in the transformer architecture the context is injected. Similar to Eq.~\ref{eq:ctx_prop}, we compute the one-sample \emph{context proportion} $r_{\text{ctx}}$, 
which measures the average attention from target queries $Q$ to context tokens $\mathcal{C}$ across all heads $H$. It quantifies how much of the attention is assigned to the context image tokens. 
As shown in Fig.~\ref{fig:person_clean}, the $r_{\text{ctx}}$ is notably high in the front-to-middle (closer to input) blocks before our attack. This indicating a strong reliance on condition-image features during these stages. 

\begin{figure*}[!t]
    \centering
    \includegraphics[width=\textwidth]{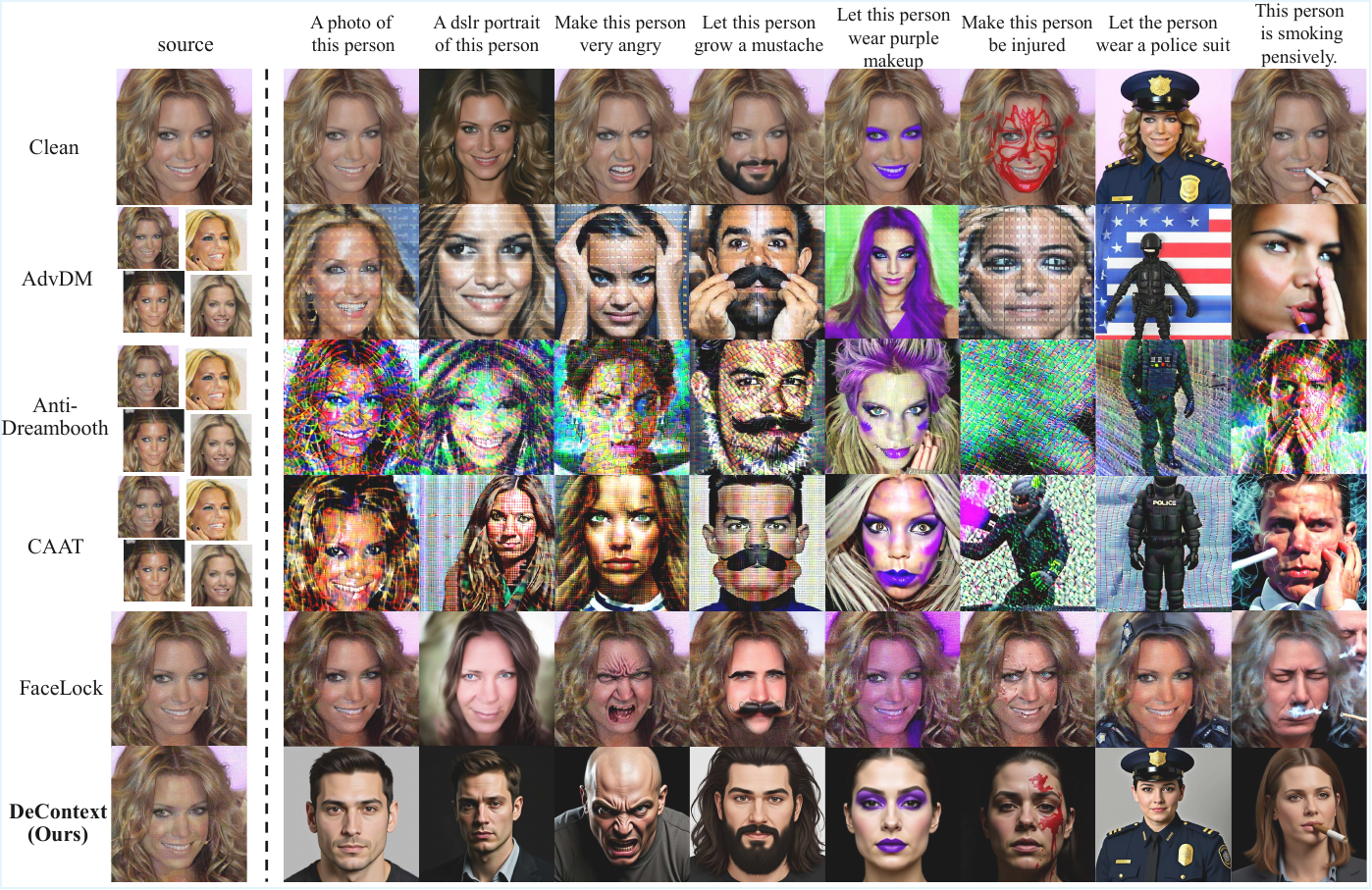}
    \vspace{-5mm}
    \caption{Defense Comparison. Previous attacks on T2I fine-tuned models and I2I models introduce visual artifacts in the generated images, whereas DeContext effectively removes context identity while preserving high visual quality.}
    \label{Fig:result_compare}
    \vspace{-3mm}
\end{figure*}

\noindent\underline{Strategy.} Therefore, we focus our suppression strategies on these specific, context-heavy blocks\footnote{Implementation details are discussed in Sec.~\ref{sec:ablation}}. As shown in Fig.~\ref{fig:person_pert}, applying this strategy significantly suppresses $r_{\text{ctx}}$ across all blocks. This confirms that our concentrated attack effectively disrupts the propagation of context information, achieving the intended detachment.

%% file: sec/5_Experiment.tex
\section{Experiments}

\subsection{Experimental Setup}
\label{sec:experimentalSetup}

\noindent\textbf{Research Goal.}~We optimize imperceptible perturbations on the context image and feed them to DiT-based image-to-image diffusion models at inference. Unlike prior works~\cite{antidreambooth,mist,caat,TargetCrossAttnAttack,DisDiff} that produce visible corruptions, our goal is to break identity preservation: generating realistic, high-quality images unrelated to the original. 


\noindent\textbf{Datasets.}~Experiments are conducted on high-quality face images derived from two datasets: VGGFace2~\cite{cao2018vggface2datasetrecognisingfaces} and CelebA-HQ~\cite{liu2015deeplearningfaceattributes}. 
For each dataset, we work with a subset containing 50 distinct identities. 
The original images are cropped and resized $512\times512$ for convenient verification.  
For T2I baselines, we select four images per identity for DreamBooth fine-tuning, following their default settings.  For I2I tasks,  the same single image per identity (one of the four used for T2I personalization) is used as context input to the diffusion model. 

\begin{table*}[t]
\centering
\setlength{\tabcolsep}{3pt} 
\renewcommand{\arraystretch}{1.0} 
\resizebox{0.88\textwidth}{!}{ 
\begin{tabular}{l l | c | c c c c c | c c c c c}
\toprule
\multirow{2}{*}{\textbf{Dataset}} & \multirow{2}{*}{\textbf{Method}} & \multirow{2}{*}{\textbf{I2I}} &
\multicolumn{5}{c|}{\textbf{``a photo of \textit{this} person''}} &
\multicolumn{5}{c}{\textbf{``a dslr portrait of \textit{this} person''}} \\
\cmidrule(lr){4-8} \cmidrule(lr){9-13}
 &  &  & \textbf{FDFR$\downarrow$} & \textbf{ISM$\downarrow$} & \textbf{CLIP-I$\downarrow$} & \textbf{BRISQUE$\downarrow$} & \textbf{FID$\downarrow$}
    & \textbf{FDFR$\downarrow$} & \textbf{ISM$\downarrow$} & \textbf{CLIP-I$\downarrow$} & \textbf{BRISQUE$\downarrow$} & \textbf{FID$\downarrow$} \\
\midrule
\multirow{7}{*}{VGGFace2}
 & Clean & \ding{51} & 0.00 & 0.78 & 0.96 & 15.78 &  200.34& 0.00 & 0.71 & 0.88 & 14.22 &  183.33\\
 \hdashline
 & Anti-DB~\cite{antidreambooth}  & \ding{55} & 0.76 & 0.21 & - & 37.33 &  421.70& 0.86 & 0.23 & - & 40.92 &  483.54\\
 & Adv-DM~\cite{advdm} & \ding{55} & 0.30  & 0.20 & - &  35.84 & 465.58 & 0.11 & \textbf{0.10} & -  & \textbf{30.37} & 280.18 \\
 & CAAT~\cite{caat} & \ding{55} & 0.80 & 0.17 & - & 39.44 & 428.80 & 0.86 & 0.17 & - & 39.11 & 429.96 \\
 & FaceLock~\cite{wang2025editawayfacestay} & \ding{51} & 0.01 & 0.19 & 0.54 & 30.75 & 233.42 & 0.01  & 0.16 & 0.59 & 40.65 & 303.10\\
 \rowcolor{gray!20} & \textbf{Diff-PGD} & \ding{51} & 0.00 & 0.60 & 0.76 & 74.19 & 253.85 & 0.00 & 0.58 & 0.77 & 45.91 & 213.57 \\
 \rowcolor{gray!20} & \textbf{DeContext} & \ding{51} & \textbf{0.00} & \textbf{0.16} & \textbf{0.50} & \textbf{23.80} & \textbf{210.98} & \textbf{0.00} & 0.23 & \textbf{0.56} & 36.83 & \textbf{207.19}  \\
\midrule
\multirow{7}{*}{CelebA-HQ}
 & Clean & \ding{51} & 0.00 & 0.79 & 0.95 & 14.20 & 130.01 & 0.00 & 0.61 & 0.82 & 12.00 & 139.22 \\
 \hdashline
 & Anti-DB~\cite{antidreambooth} & \ding{55} & 0.73 & 0.36 & - & 38.83 &  395.27& 0.74 & 0.27 & - & 38.99 &  424.60\\
 & AdvDM~\cite{advdm} & \ding{55} & 0.07 & 0.35 &  - & 22.67 & 280.91 & 0.03 & 0.22 & - & \textbf{28.79} & 231.49 \\
 & CAAT~\cite{caat} & \ding{55} & 0.60 & 0.32 & - & 46.01 & 413.68  & 0.78 & 0.26 & - & 40.16 & 370.56 \\
 & FaceLock~\cite{wang2025editawayfacestay} & \ding{51} & 0.00 & 0.51 &  0.73 &29.95 &  \textbf{200.87} & 0.00 & 0.40 & 0.65 & 40.96 &  231.15\\
 \rowcolor{gray!20} & \textbf{Diff-PGD} & \ding{51} & 0.01 & 0.68 & 0.80 & 74.30 & 235.89 & 0.00 & 0.59 & 0.79 & 38.96 & 230.72  \\
 \rowcolor{gray!20} & \textbf{DeContext} & \ding{51} & \textbf{0.00} & \textbf{0.12} & \textbf{0.51} & \textbf{20.56} & 229.68 & \textbf{0.00} & \textbf{0.20} & \textbf{0.58} & 39.17 & \textbf{209.37} \\
\bottomrule
\end{tabular}
}
\vspace{-3mm}
\caption{Quantitative results on VGGFace2 and CelebA-HQ datasets.}
\vspace{-3mm}
\label{tab:defense_comparison}
\end{table*}

\begin{table}[h]
\centering
\setlength{\tabcolsep}{2.5pt}
\renewcommand{\arraystretch}{1.0}
\footnotesize
\begin{tabular}{l *{5}{c}}
\toprule
\multirow{2}{*}{Settings} & \multicolumn{5}{c}{\textbf{``Make this person very angry.''}} \\
\cmidrule(lr){2-6}
& \textbf{ISM}$\downarrow$ & \textbf{CLIP-I}$\downarrow$ & FID$\downarrow$ & BRISQUE$\downarrow$ & SER-FIQ$\uparrow$ \\
\midrule
clean & 0.60 & 0.88 & 139.41 & 13.67 & 0.78 \\
DeContext & 0.07 & 0.49 & 252.44 & 33.17 & 0.79\\
\midrule
\multirow{2}{*}{Settings} & \multicolumn{5}{c}{\textbf{``This person is wearing a smoky eye makeup.''}} \\
\cmidrule(lr){2-6}
& \textbf{ISM}$\downarrow$ & \textbf{CLIP-I}$\downarrow$ & FID$\downarrow$ & BRISQUE$\downarrow$ & SER-FIQ$\uparrow$ \\
\midrule
clean & 0.75 & 0.91 & 141.28  & 13.77 & 0.74\\
DeContext & 0.21 & 0.58 & 202.77 & 14.10 & 0.60\\
\midrule
\multirow{2}{*}{Settings} & \multicolumn{5}{c}{\textbf{``This person is looking at the mirror.''}} \\
\cmidrule(lr){2-6}
& \textbf{ISM}$\downarrow$ & \textbf{CLIP-I}$\downarrow$ & FID$\downarrow$ & BRISQUE$\downarrow$ & SER-FIQ$\uparrow$ \\
\midrule
clean & 0.72 & 0.85 & 190.73  & 13.12 & 0.72\\
DeContext & 0.36 & 0.61 & 241.49 & 10.55 & 0.67\\
\midrule
\multirow{2}{*}{Settings} & \multicolumn{5}{c}{\textbf{``Add glasses to this person.''}} \\
\cmidrule(lr){2-6}
& \textbf{ISM}$\downarrow$ & \textbf{CLIP-I}$\downarrow$ & FID$\downarrow$ & BRISQUE$\downarrow$ & SER-FIQ$\uparrow$ \\
\midrule
clean & 0.60 & 0.88 & 179.71 & 15.00 & 0.73 \\
DeContext & 0.14 & 0.57 & 212.45 & 16.24 & 0.76 \\
\bottomrule
\end{tabular}
\vspace{-3mm}
\caption{Quantitative results under different prompts.}
\vspace{-3mm}
\label{tab:multi_prompts}
\end{table}
\noindent\textbf{Target Models.}~Our primary evaluation target is FLUX.1-Kontext-dev~\cite{flux1kontext}, a state-of-the-art DiT-based in-context editing model. We also evaluate generalization on Step1X-Edit~\cite{step1xedit} (Sec~\ref{sec:step1x-edit}).

\noindent\textbf{Baselines.} 
Since no prior work directly addresses DiT-based I2I editing defense, we adapt existing methods:
\begin{itemize}[leftmargin=*,itemsep=0pt]
\item \textit{T2I personalization defenses: }Anti-DB~\cite{antidreambooth}, AdvDM~\cite{advdm}, and CAAT~\cite{caat} applied to DreamBooth.
\item \textit{I2I defense:} FaceLock~\cite{wang2025editawayfacestay} for InstructPix2Pix~\cite{instructP2P}.
\item \textit{Naive baseline: }Diff-PGD maximizing reconstruction loss. The working principle is discussed in Sec.~\ref{sec:observation}.
\end{itemize}

\noindent\textbf{Implementation Details.}
We set the step size to \(\alpha = 0.005\) and the noise budget to \(\eta = 0.1\), running a total of 800 optimization steps. At each training iteration, we randomly sample a diffusion timestep $t$ from the interval $\{980, \dots, 1000\}$, and a prompt from a fixed pool of 60 prompts (complete list provided in the Appendix \ref{promptpool}). 
Experiments are conducted on a single NVIDIA A800 80 GB GPU.
Unless otherwise specified, these settings remain consistent across all experiments.

\begin{table*}[!t]
\centering
\renewcommand{\arraystretch}{1.1}
\setlength{\tabcolsep}{3pt}
\resizebox{0.9\textwidth}{!}{%
\begin{tabular}{l|l|ccccc|ccccc}
\toprule
\multicolumn{1}{c|}{} & \multicolumn{1}{c|}{} & \multicolumn{5}{c|}{\textbf{``a photo of this person''}} & \multicolumn{5}{c}{\textbf{``a dslr portrait of this person''}} \\ 
\cline{3-12} 
\multicolumn{1}{c|}{\multirow{-2}{*}{Dataset}} & 
\multicolumn{1}{c|}{\multirow{-2}{*}{Method}} &
\textbf{ISM}$\downarrow$ & \textbf{CLIP-I}$\downarrow$ & FID$\downarrow$ & BRISQUE$\downarrow$ & SER-FIQ$\uparrow$ &
\textbf{ISM}$\downarrow$ & \textbf{CLIP-I}$\downarrow$ & FID$\downarrow$ & BRISQUE$\downarrow$ & SER-FIQ$\uparrow$ \\ 
\hline
 & Clean & 0.62 & 0.92 & 125.64 & 10.82 & 0.76 & 0.67 & 0.90 & 166.49 & 8.51 & 0.73 \\ 
 \rowcolor{gray!20}
 & \textbf{DeContext} & 0.13 & 0.56 & 160.87  & 18.89 & 0.81  & 0.16 & 0.60 &  183.37 & 13.84 & 0.77\\ 
\cline{2-12} 
CelebA-HQ &  & 
\multicolumn{5}{c|}{\textbf{``This person is wearing a smoky eye makeup.''}} & 
\multicolumn{5}{c}{\textbf{``Give this person a beard.''}} \\ 
\cline{3-12} 
 & \multirow{-2}{*}{Method} & 
\textbf{ISM}$\downarrow$ & \textbf{CLIP-I}$\downarrow$ & FID$\downarrow$ & BRISQUE$\downarrow$ & SER-FIQ$\uparrow$ &
\textbf{ISM}$\downarrow$ & \textbf{CLIP-I}$\downarrow$ & FID$\downarrow$ & BRISQUE$\downarrow$ & SER-FIQ$\uparrow$ \\ 
\cline{2-12} 
 & Clean & 0.67 & 0.92 & 134.42 & 8.87 & 0.74 & 0.57 & 0.76 & 161.25 & 5.28 &  0.67 \\ 
 \rowcolor{gray!20}
 & \textbf{DeContext} & 0.17 &  0.59 &  161.89 & 15.45 & 0.72 & 0.08 & 0.50 & 236.15 & 15.69 & 0.70 \\ 
\bottomrule
\end{tabular}%
}
\vspace{-3mm}
\caption{Evaluations on Step1X-Edit.}
\vspace{-3mm}
\label{tab:step1x-edit}
\end{table*}

\begin{table}[!t]
\centering
\renewcommand{\arraystretch}{1.1}
\resizebox{0.5\textwidth}{!}{
\begin{tabular}{llccccc}
\toprule
\multirow{2}{*}{\textbf{Dataset}} & 
\multirow{2}{*}{$\eta$} &
\multicolumn{5}{c}{\textbf{“a photo of \textit{this} person”}} \\
\cmidrule(lr){3-7}
 & & \textbf{ISM}$\downarrow$ & \textbf{CLIP-I}$\downarrow$ & SER-FQA$\uparrow$ & BRISQUE$\downarrow$ & FID$\downarrow$ \\
\midrule
\multirow{4}{*}{\textbf{VGGFace2}} 
 & 0.00  & 0.78 & 0.96 & 0.76 & 15.78 & 200.34 \\
 \hdashline
 & 0.05  & 0.40 & 0.67 & \textbf{0.71} & \textbf{20.05} & \textbf{197.86} \\
 & 0.10* & 0.16 & 0.51 & 0.65 & 23.80 & 210.98 \\
 & 0.15  & \textbf{0.14} & \textbf{0.50} & 0.65 & 24.17 & 213.54 \\
\midrule
\multirow{4}{*}{\textbf{CelebA-HQ}} 
 & 0.00  & 0.79 & 0.95 & 0.76 & 14.20 & 130.01 \\
 \hdashline
 & 0.05  & 0.36 & 0.66 & \textbf{0.69} & \textbf{15.38} & \textbf{177.95} \\
 & 0.10* & 0.12 & 0.51 & 0.64 & 20.56 & 229.68 \\
 & 0.15  & \textbf{0.11} & \textbf{0.50} & 0.64 & 22.87 & 228.72 \\
\bottomrule
\end{tabular}%
}
\vspace{-3mm}
\caption{Ablation on attack budgets $\eta$. (* denotes default)}
\vspace{-2mm}
\label{tab:budgets}
\end{table}

\begin{table}[!t]
\centering
\setlength{\tabcolsep}{6pt}
\renewcommand{\arraystretch}{1}
\footnotesize
\begin{tabular}{l c c c}
\toprule
\multicolumn{1}{c}{} & \multicolumn{3}{c}{\textbf{Perturbed Blocks}} \\[-2pt]
\cmidrule(lr){2-4}
\textbf{Metrics} & \textbf{Double All} & \textbf{Single 0--25*} & \textbf{Single 12--37} \\
\midrule
ISM $\downarrow$ & 0.52 & 0.13 & 0.20 \\
CLIP-I $\downarrow$ & 0.73 & 0.49 & 0.57 \\
\bottomrule
\end{tabular}
\vspace{-3mm}
\caption{Ablation on different perturbed blocks.}
\vspace{-3mm}
\label{tab:perturbed_blocks}
\end{table}

\noindent\textbf{Evaluation metrics.}
Following prior text-to-image personalization and editing works, we use the same evaluation protocol for fair baseline comparison. We evaluate the output images from two aspects: \textit{whether the identity is altered} and \textit{how well the image quality is preserved}.




\begin{itemize}
    \item \textbf{Identity Protection.}
We use \textbf{\textit{Face Detection Failure Rate (FDFR)}} to measure the percentage of samples where RetinaFace~\cite{retinaface} fails to detect a face. Once detected, we use\textbf{ \textit{Identity Score Matching (ISM)}} to compute the distance between ArcFace~\cite{ArcFace} embeddings, with smaller values indicating greater identity change.  For I2I tasks, we additionally compute \textit{\textbf{CLIP Image Similarity (CLIP-I)}}, which measures semantic similarity between source and generated images.
    \item \textbf{Image Quality.}
measured by \textit{\textbf{SER-FQA}}~\cite{SER-FIQ} (face-specific), \textit{\textbf{BRISQUE}}~\cite{BRISQUE} (perceptual) and\textit{ \textbf{FID}} (statistical realism).
Since we aim to remove identity cues while keeping realism, higher SER-FQA, lower BRISQUE and lower FID denote better results.
\end{itemize}
While we mainly evaluate on faces as the most representative and safety-critical example, we additionally include object-level results in the Appendix~\ref{visualize}.

\subsection{Main Results}
\noindent\textbf{Comparison with protection baselines.} 
To evaluate whether DeContext provides stronger protection for user image privacy, we compare it against prior UNet-based defenses under the same setup.
Clean experiments are conducted on Flux Kontext and we rerun all baselines under noise budget \(\eta = 0.1\) for a fair comparison. 
For T2I personalization on DreamBooth, we follow Anti-DreamBooth~\cite{antidreambooth} prompts ``a photo of sks person'' and ``a dslr portrait of sks person'', where \textit{sks} is the keyword.  
For I2I tasks, prompts are adapted to ``a photo of this person'' and ``a dslr portrait of this person''.  
These prompts are excluded in our 60-prompt training pool, to show that DeContext generalizes to test prompts.  
Each identity–prompt pair generates 30 samples.

As shown in Tab.~\ref{tab:defense_comparison},  DeContext generally outperforms prior UNet-based defenses.
In terms of identity removal, DeContext detects more faces while more effectively eliminating identifiable information. On CelebA-HQ using the prompt “a photo of this person”, it achieves an ISM of 0.12, markedly better than the best baseline (0.32). CLIP-I scores also remain consistently lower than those of the state-of-the-art I2I defense FaceLock~\cite{wang2025editawayfacestay}.
For image quality, DeContext also achieves lower Brisque and FID, producing fewer artifacts and more natural outputs. Note that, the Diff-PGD baseline performs poorly, with notably high ISM and CLIP-I scores.
As shown in Fig.~\ref{Fig:result_compare}, prior T2I defenses (Anti-DreamBooth, AdvDM, CAAT) introduce severe distortions with colorful noise and corrupted textures, while FaceLock produces color shifts and sometimes fails to fully remove identity. DeContext, in contrast, consistently generates realistic outputs that are completely unrelated to the source identity across all editing instructions, achieving the best protection-quality trade-off.

\noindent\textbf{Generalization to multiple face editing prompts. } We further evaluate the robustness of DeContext under diverse editing prompts. Specifically, we test it on CelebA-HQ using four additional in-context editing instructions. For each prompt, we sample 30 distinct identities and generate 10 images per identity.

With quantitative results in Tab.~\ref{tab:multi_prompts}, identity metrics show notable reductions across all prompts: ISM and CLIP-I decrease by 73\% and 36\% on average. Image quality remains stable, with BRISQUE and SER-FIQ varying under 10\% for most cases. Since the real FID embeddings are derived from clean faces, they align closely with clean outputs but not with perturbed ones, leading to relatively higher FID. Qualitative results in Fig.~\ref{Fig:result_compare} visually show that other baselines tend to preserve identity to some extent and often produce visible artifacts. In contrast, DeContext achieves a clear trade-off between identity removal and overall visual quality.

\subsection{Ablation studies}
\label{sec:ablation}

To evaluate the impact of the attack budget and our block-wise concentrated detachment design, we conduct two ablation experiments following the setup in Sec.~\ref{sec:experimentalSetup}.

\noindent\textbf{Ablation on attack budget.} We examine how performance varies with the attack’s norm ball size.
As shown in Tab.~\ref{tab:budgets}, a larger budget $\eta$ improves identity detachment but introduces more visual artifacts and slightly reduces image quality. We use $\eta=0.1$ as the default.

\noindent\textbf{Ablation on perturbed blocks.}
Flux-Kontext consists of 19 double blocks followed by 38 single blocks. To validate our finding and layer selection in Sec.~\ref{sec:context_propagation}, we test three settings: attacking all double blocks, the first 25 single blocks, and the last 25 single blocks (Tab.~\ref{tab:perturbed_blocks}). Consistent with our block-wise analysis, early-to-mid single blocks play a key role in context propagation. Perturbing only these blocks achieves the best trade-off between identity detachment and efficiency.


\subsection{Extension Results on Step1X-Edit}
\label{sec:step1x-edit}
Step1X-Edit~\cite{step1xedit} is another DiT-based model designed for image editing. To assess the generalization capability of our proposed attack, we directly apply DeContext to Step1X-Edit using the same experimental setup as described for FLUX.1-Kontext in Section~\ref{sec:experimentalSetup}. Specifically, we evaluate on the CelebA-HQ~\cite{liu2015deeplearningfaceattributes} dataset,  generating 50 images per identity-prompt pair.

\begin{figure}[t]
    \centering
    \includegraphics[width=0.45\textwidth]{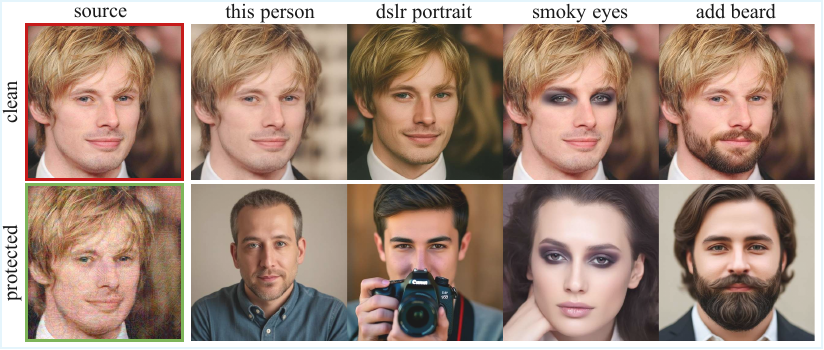}
    \caption{Qualitative results on Step1X-Edit.}
    \vspace{-3mm}
    \label{fig:settozero}
\end{figure}

As shown in Fig.~\ref{fig:settozero}, DeContext reliably avoids identity transfer while producing realistic results across a wide range of editing directions from neutral prompts (``dslr portrait") to specific appearance changes (``smoky eyes", ``add beard"). 
Quantitatively, as summarized in Tab.~\ref{tab:step1x-edit}, DeContext achieves comparable identity removal and image quality performance to that observed on FLUX.1-Kontext, with a notable average ISM reduction of more than 80\%. These results demonstrate that DeContext maintains its effectiveness across different DiT-based image-to-image architectures. 

%% file: sec/6_conclusion.tex
\section{Conclusion and Future Work}

In this paper, we present DeContext, the first defense for DiT-based in-context image editing. DeContext protects user images by disrupting the propagation of context through multi-modal attention. Experiments on FLUX.1-Kontext and Step1X-Edit demonstrate that DeContext effectively detaches the contextual information from the input while preserving high quality. In future work, we aim to improve defense efficiency on large models and enhance robustness and transferability in black-box settings for real-world protection.

%% file: sec/X_suppl.tex
\maketitle
\appendix

In this supplement, we detail the design, implementation, extension experiments and analysis of DeContext: pseudocode in Sec.\ref{imple details}, experiment on items in Sec.\ref{items exp}, user perference study in Sec.\ref{user study}, a discussion of limitation in Sec.\ref{analysis}, and visuals in Sec.\ref{visualize}.



\section{Implementation Details}
\label{imple details}
\subsection{Diff-PGD Baseline}
\label{alogorithms}
We first implement an intuitive adversarial baseline \textbf{Diff-PGD}, where perturbations on the condition image $x^{\mathrm{adv}} = x + \delta$ are optimized to maximize the \emph{flow matching} loss. The goal is to prevent the model from generating the desired output when conditioned on \(x^{\mathrm{adv}}\).

Specifically, at each sampled 
diffusion timestep \(t\), the I2I model \(\mathcal{G}\) predicts the velocity term:
\[
\hat{v}_t = \mathcal{G}\!\left(z_t,\ \mathcal{E}_{\mathrm{vae}}(x^{\mathrm{adv}}),\ p,\ t\right),
\]
where \(z_t\) denotes the noisy latent at timestep \(t\), 
\(\mathcal{E}_{\mathrm{vae}}(\cdot)\) encodes the perturbed condition image into latent 
space, and \(p\) is a randomly sampled prompt. The ground truth target latent is 
given by \(z_t^\ast\), and the reconstruction-driven objective encourages the 
predicted velocity to deviate from this target to achieve protection:
\[
\mathcal{L}_{\mathrm{recon}}
= \mathbb{E}_{t,p,z}\!\left[\left\| \hat{v}_t - (z_t - z_t^\ast) \right\|_2^2\right].
\]
The perturbation \(\delta\) is optimized via projected gradient ascent under an 
\(\ell_\infty\) constraint, as shown in Algorithm~\ref{alg:flux_recon}.
\begin{algorithm}[htbp]  
\caption{Diff-PGD: Reconstruction Attack}
\label{alg:flux_recon}
\begin{algorithmic}[1]
\Require Condition image $x$, prompt pool $\mathcal{P}$, target latents $\{z_i^\ast\}$, diffusion scheduler $\mathcal{S}$, step size $\alpha$, perturbation bound $\epsilon$, total steps $N$.
\Ensure Perturbed image $x^{\text{adv}}$.
\State Initialize $\delta \gets 0$, \quad $x^{\text{adv}} \gets x$
\For{$i = 1$ to $N$}
    \State Sample $p_i \sim \mathcal{P}$, $t \sim \mathcal{U}(t, T)$, $z \sim \mathcal{N}(0,I)$
\State $\sigma_t \gets \mathcal{S}(t)$, \quad $z_t  \gets (1-\sigma_t)z^\ast + \sigma_t n$
    \State $c^{\text{adv}} \gets \mathcal{E}_{\text{vae}}(x^{\text{adv}})$
    \State Forward through I2I diffusion model $\mathcal{G}$
    \State $\hat{v} \gets \mathcal{G}(\tilde{z}, c^{\text{adv}}, p_t, t)$
    \State $\mathcal{L} \gets \|\hat{v} - (z_t - z^\ast)\|_2^2$
    \State $g \gets \nabla_{x^{\text{adv}}}\mathcal{L}$
    \State $x^{\text{adv}} \gets x^{\text{adv}} + \alpha\,\text{sign}(g)$
    \State $x^{\text{adv}} \gets \text{clip}(x^{\text{adv}},\, x-\epsilon,\, x+\epsilon)$
\EndFor
\State \Return $x^{\text{adv}}$
\end{algorithmic}
\end{algorithm}

Regardless of whether we target the early denoising stages (higher timesteps), which are critical for propagating context information, Diff-PGD fails to produce meaningful adversarial effects in DiT-based image-to-image models.

\subsection{DeContext}

We then implement DeContext to intentionally suppress the influence of context images by weakening their cross-attention contributions. The detailed attack strategy is introduced in Algorithm~\ref{alg:flux_attack}.

\begin{algorithm}[H]  
\caption{DeContext: Attention-aware Attack.}
\label{alg:flux_attack}
\begin{algorithmic}[1]
\Require Condition image $x$, prompt pool $\mathcal{P}$, diffusion scheduler $\mathcal{S}$ with $T$ timesteps, step size $\alpha$, perturbation bound $\epsilon$, total steps $N$.
\Ensure Perturbed image $x^{\text{adv}}$.
\State Initialize $\delta \gets 0$, \quad $x^{\text{adv}} \gets x$
\For{$i = 1$ to $N$}
    \State Sample $p_i \sim \mathcal{P}$, $t_i \sim \mathcal{U}(t_{\text{high}}, T)$, $z \sim \mathcal{N}(0,I)$
    \State $z_t \gets \mathcal{S}(z,t_i)$, \quad $c^{\text{adv}} \gets \mathcal{E}_{\text{vae}}(x^{\text{adv}})$
    \State Single blocks forward through Transformer $\mathcal{T}$
    \State $(q,k)\gets \mathcal{T}(z_t, c^{\text{adv}}, p_i,t)$
    \State $A_{q,:}\gets\operatorname{Softmax}\!\big((qk^\top)/\sqrt{d}\big)$ \Comment{target querys}
    \State $A_{\text{ctx}}=\{A_{qk}\mid k\in\mathcal{C}\}$ \Comment{context keys}
    \State $r_{\text{ctx}}\gets\frac{1}{H B|Q|}\sum_{h,b}\sum_{q\in Q}\sum_{k\in\mathcal{C}}A^{(h,b)}_{qk}$ 
    
    \Comment{mean attention over heads, blocks, queries}
    \State $\mathcal{L} \gets 1 - r_{\text{ctx}}$, \quad $g \gets \nabla_{x^{\text{adv}}}\mathcal{L}$
    \State $x^{\text{adv}} \gets \text{clip}(x^{\text{adv}} + \alpha \cdot \text{sign}(g),\, x - \epsilon,\, x + \epsilon)$
\EndFor
\State \Return $x^{\text{adv}}$
\end{algorithmic}
\end{algorithm}

For the concentrated attack setting, we target single blocks from 0 to 25 and sample timesteps in the range of 980 to 1000 by default.
Although later layers are not directly optimized, they may be affected indirectly by the disrupted signals coming from the attacked blocks.
Noticed that, at the selected high-noise steps, we approximate the target image with random Gaussian noise $z \sim \mathcal N (0, I)$, avoiding the need for paired target images corresponding to each prompt, making the optimization both simpler and more flexible.

\subsection{Prompt Pool}
\label{promptpool}

To promote prompt-agnostic generalization during adversarial optimization, we construct a diverse prompt pool $\mathcal{P}$ containing 60 natural-language instructions categorized into six groups: facial expressions, accessory addition, posture and movement, style changes, scene changes, and combined edits. 

\paragraph{(1) Facial Expressions}
\begin{itemize}[leftmargin=2em]
    \item Make this person smile happily.
    \item Make this person look angry.
    \item Make this person look surprised.
    \item Make this person look sad.
    \item Give this person a serious expression.
    \item Make this person laugh joyfully.
    \item Make this person look worried.
    \item Make this person look excited.
    \item Make this person look tired.
    \item Make this person look confused.
\end{itemize}

\paragraph{(2) Accessory Addition}
\begin{itemize}[leftmargin=2em]
    \item Add glasses to this person.
    \item Add sunglasses to this person.
    \item Add a hat to this person.
    \item Add a scarf to this person.
    \item Add a necklace to this person.
    \item Add earrings to this person.
    \item Add a cap to this person.
    \item Add a headband to this person.
    \item Add a watch to this person.
    \item Add a tie to this person.
\end{itemize}

\paragraph{(3) Posture and Movement}
\begin{itemize}[leftmargin=2em]
    \item Make this person wave their hand.
    \item Make this person cross their arms.
    \item Make this person point forward.
    \item Make this person give a thumbs up.
    \item Make this person put hands on hips.
    \item Make this person cover their mouth.
    \item Make this person hold up a peace sign.
    \item Make this person shrug their shoulders.
    \item Make this person touch their chin.
    \item Make this person salute.
\end{itemize}

\paragraph{(4) Style Changes}
\begin{itemize}[leftmargin=2em]
    \item Make this person look older.
    \item Make this person look younger.
    \item Add a beard to this person.
    \item Add a mustache to this person.
    \item Make this person's hair longer.
    \item Make this person's hair shorter.
    \item Add makeup to this person.
    \item Add freckles to this person.
    \item Make this person's skin tanned.
    \item Add stubble to this person.
\end{itemize}

\paragraph{(5) Scene Changes}
\begin{itemize}[leftmargin=2em]
    \item Place this person on a tropical beach with palm trees and ocean waves.
    \item Put this person in a snowy mountain landscape with pine trees.
    \item Move this person to a bustling city street with skyscrapers.
    \item Put this person in a beach setting.
    \item Put this person in a snowy mountain.
    \item Put this person in a city street.
    \item Put this person in a forest.
    \item Put this person in a coffee shop.
    \item Put this person in a library.
    \item Put this person in a park.
\end{itemize}

\paragraph{(6) Combined Prompts}
\begin{itemize}[leftmargin=2em]
    \item Make this person smile and wave.
    \item Add glasses and make this person look serious.
    \item Make this person look older with a beard.
    \item Add a hat and make this person point forward.
    \item Make this person look surprised with hands on face.
    \item Add sunglasses and make this person give a thumbs up.
    \item Make this person look younger and smile happily.
    \item Add a scarf and make this person cross their arms.
    \item Make this person look tired and rub their eyes.
    \item Add earrings and make this person look confident.
\end{itemize}

\section{Extension Experiments on Items}
\label{items exp}
In the main paper, we primarily focus on the domain of human face generation to mitigate potential misuse.
To assess the generalizability of DeContext beyond human portraits, we also conduct experiments on diverse contextual images featuring various object items.

\noindent\textbf{Experiment Settings.} We select 50 item images from the OminiControl Subject200K dataset\cite{ominicontrol}, each resized to 512$\times$512 resolution. We follow the training settings used for human portraits and applied them to items, modifying only the prompt pool to better adapt them to item-related content.
For the clean and perturbed version of each item, 10 outputs are generated per image-prompt pair with different seeds.

\noindent\textbf{Evaluation Metrics.} We employ the following metrics to quantitatively evaluate the results:
\begin{itemize}
    \item \textbf{DINO}\cite{dino}: Extracts image features to evaluate semantic similarity of items, focusing on shape and structure regardless of background.
    \item \textbf{CLIP-Image}\cite{clip}: Uses image embeddings to measure visual-semantic consistency, capturing category and style-level similarity.
    \item \textbf{SSIM}\cite{SSIM}: Computes structural similarity to assess pixel-wise fidelity between generated and reference images.
\end{itemize}

\noindent\textbf{Results.} We evaluate item-level context detachment using DINO, CLIP-Image, and SSIM by comparing images generated from perturbed versus clean contexts. As shown in Tab.~\ref{tab:item_multi_prompts}, across six prompts, the average similarity scores decrease by 58\%, 25\%, and 64\%, respectively, indicating substantial divergence and successful contextual decoupling.
Visual examples in Fig.~\ref{Fig:item_results} further confirm that our method offers a reasonable defense against image misuse.

\begin{table}[H]
\centering
\setlength{\tabcolsep}{7pt}
\renewcommand{\arraystretch}{1.1}
\footnotesize
\begin{tabular}{l ccc}
\toprule
\multirow{2}{*}{Settings} & \multicolumn{3}{c}{\textbf{``A photo of this item.''}} \\
\cmidrule(lr){2-4}
& \textbf{DINO}$\downarrow$ & \textbf{CLIP-I}$\downarrow$ & \textbf{SSIM}$\downarrow$ \\
\midrule
clean & 0.97 &0.98 & 0.61\\
DeContext & 0.36 &0.65 & 0.24\\
\midrule
\multirow{2}{*}{Settings} & \multicolumn{3}{c}{\textbf{``A top-up view of this item.''}} \\
\cmidrule(lr){2-4}
& \textbf{DINO}$\downarrow$ & \textbf{CLIP-I}$\downarrow$ & \textbf{SSIM}$\downarrow$ \\
\midrule
clean & 0.78 & 0.87 & 0.22\\
DeContext &0.32 & 0.66 & 0.14\\
\midrule
\multirow{2}{*}{Settings} & \multicolumn{3}{c}{\textbf{``Make this item look worn and old.''}} \\
\cmidrule(lr){2-4}
& \textbf{DINO}$\downarrow$ & \textbf{CLIP-I}$\downarrow$ & \textbf{SSIM}$\downarrow$ \\
\midrule
clean & 0.90 & 0.93 & 0.47\\
DeContext & 0.28 & 0.59 & 0.09\\
\midrule
\multirow{2}{*}{Settings} & \multicolumn{3}{c}{\textbf{``Illuminate this item from left.''}} \\
\cmidrule(lr){2-4}
& \textbf{DINO}$\downarrow$ & \textbf{CLIP-I}$\downarrow$ & \textbf{SSIM}$\downarrow$ \\
\midrule
clean & 0.98 & 0.97 & 0.85\\
DeContext & 0.53& 0.63 & 0.17\\
\midrule
\multirow{2}{*}{Settings} & \multicolumn{3}{c}{\textbf{``Change the color of this item to blue.''}} \\
\cmidrule(lr){2-4}
& \textbf{DINO}$\downarrow$ & \textbf{CLIP-I}$\downarrow$ & \textbf{SSIM}$\downarrow$ \\
\midrule
clean &0.94 & 0.94 & 0.79\\
DeContext &0.35 &0.65 & 0.28\\
\midrule
\multirow{2}{*}{Settings} & \multicolumn{3}{c}{\textbf{``Place this item onto a mirror.''}} \\
\cmidrule(lr){2-4}
& \textbf{DINO}$\downarrow$ & \textbf{CLIP-I}$\downarrow$ & \textbf{SSIM}$\downarrow$ \\
\midrule
clean &0.87 & 0.91 & 0.70\\
DeContext &  0.45 & 0.68 & 0.25\\
\bottomrule
\end{tabular}
\caption{Quantitative results for items under different prompts.}
\label{tab:item_multi_prompts}
\end{table}


\section{User Study}
\label{user study}

To evaluate the subjective effectiveness of our defense method, we conducted a controlled user study with 20 participants. We compared \textbf{DeContext} against four state-of-the-art image protection methods: Anti-Dreambooth~\cite{antidreambooth}, CAAT~\cite{caat}, AdvDM~\cite{mist}, and FaceLock~\cite{wang2025editawayfacestay}.

Each participant was shown 8 generated images per method, from the same input prompt and protected image. They were asked to rank them along four dimensions: (1) \textit{Identity Detachment} (how well personal identity is obscured), (2) \textit{Prompt Adherence} (how faithfully the output reflects the input prompt), (3) \textit{Image Quality} (visual realism and artifact suppression), and (4) \textit{Overall Protection Preference} (overall preference for privacy protection).

\begin{figure}[H]
    \centering
    \includegraphics[width=0.45\textwidth]{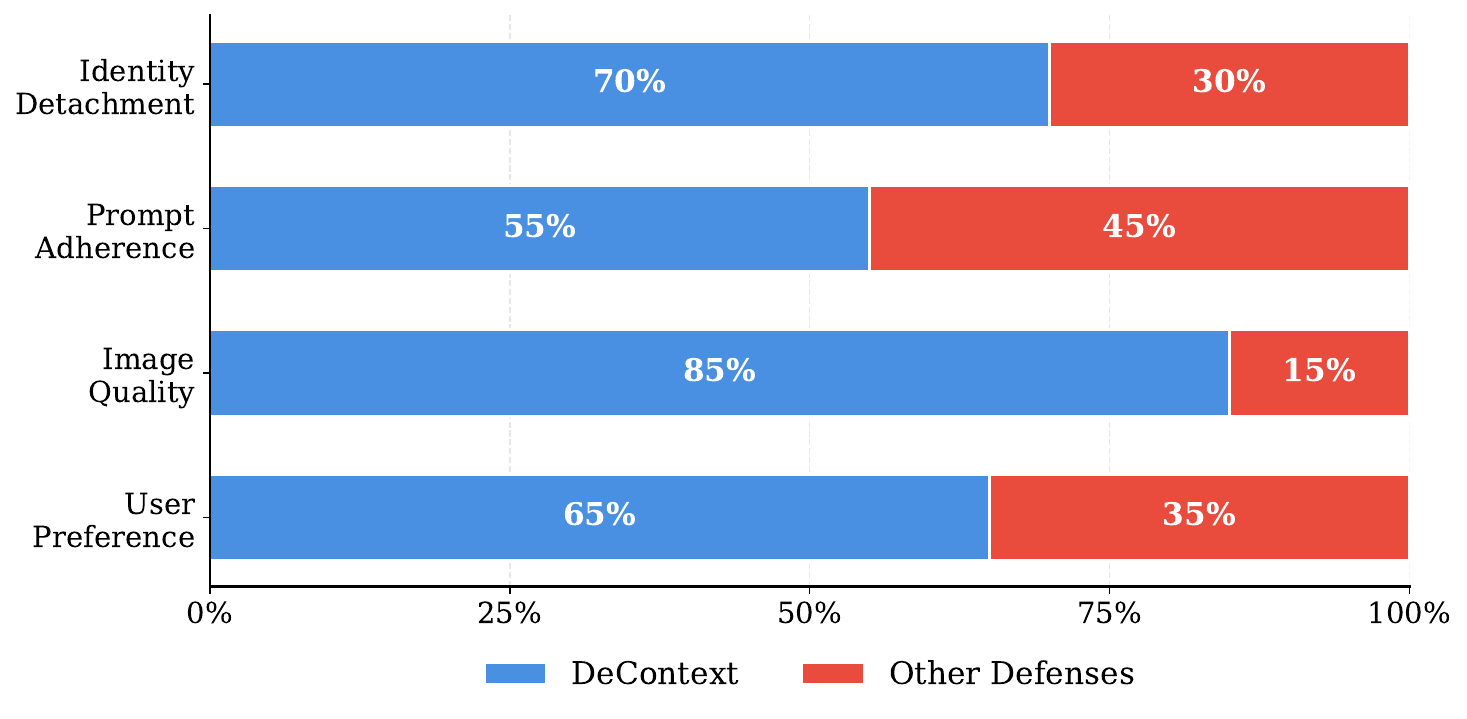}
    \caption{User Study.}
    \label{fig:user study}
\end{figure}

As shown in Fig.~\ref{fig:user study}, DeContext received the highest number of top-ranked selections. Notably, its total number of first-place rankings exceeded the combined first-place selections of all four baselines. It demonstrates a strong and consistent user preference for DeContext.




\section{Discussion and Future Work}
\label{analysis}

\noindent\textbf{Failure Case on Complex Scenes}.  
DeContext works well in simple image edits where the context image plays a dominant role. However, its effectiveness diminishes under complex scene modifications driven by language.

\begin{figure}[H]
    \centering
    \includegraphics[width=0.4\textwidth]{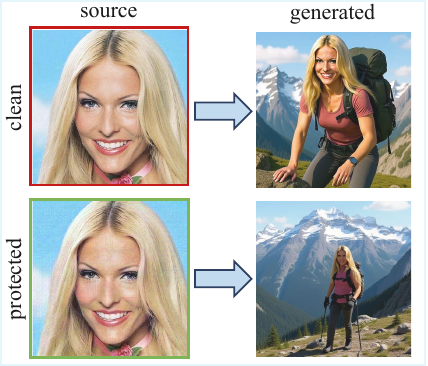}
    \caption{Failure Case with the prompt ``Transform this image to mountain hiking scene, change the person's pose to climbing stance, add backpack and snow peaks.''}
    \label{fig:failure_cases}
\end{figure}

As shown in Fig.~\ref{fig:failure_cases}, when the prompt demands a substantial transformation of the scene, the model largely disregards the original visual context, even in the absence of any defense mechanism. In such cases, generation is primarily governed by the text prompt rather than the input image. Because the model already allocates minimal attention to the visual context, further weakening cross-attention via DeContext yields only minor effects (e.g., slight resizing of a person), leaving object identity and overall scene composition largely intact. This demonstrates that strong textual guidance can override contextual signals, thereby limiting the impact of uniform context detachment.

\noindent\textbf{Future Direction}. Given these limitations, a promising avenue for future work is to shift from uniform suppression of contextual information toward selective attention reduction. By targeting the reduction more precisely at sensitive elements, the method may better limit their influence, especially in scenarios where the prompt requires significant scene alterations.

\section{Additional Qualitative Results}
\label{visualize}
\begin{figure*}[!t]
    \centering
    \includegraphics[width=0.9\textwidth]{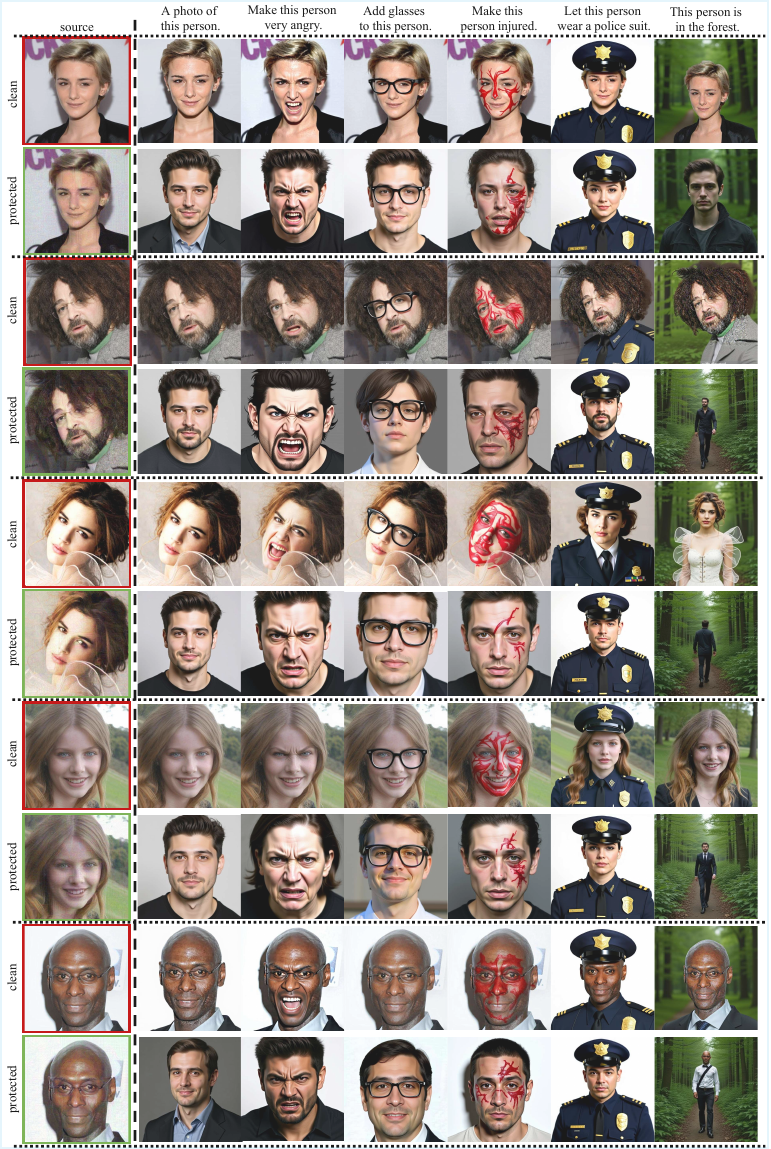}
    \caption{Defense on human portraits.}
    \label{Fig:human_results}
\end{figure*}

\begin{figure*}[!t]
    \centering
    \includegraphics[width=0.9\textwidth]{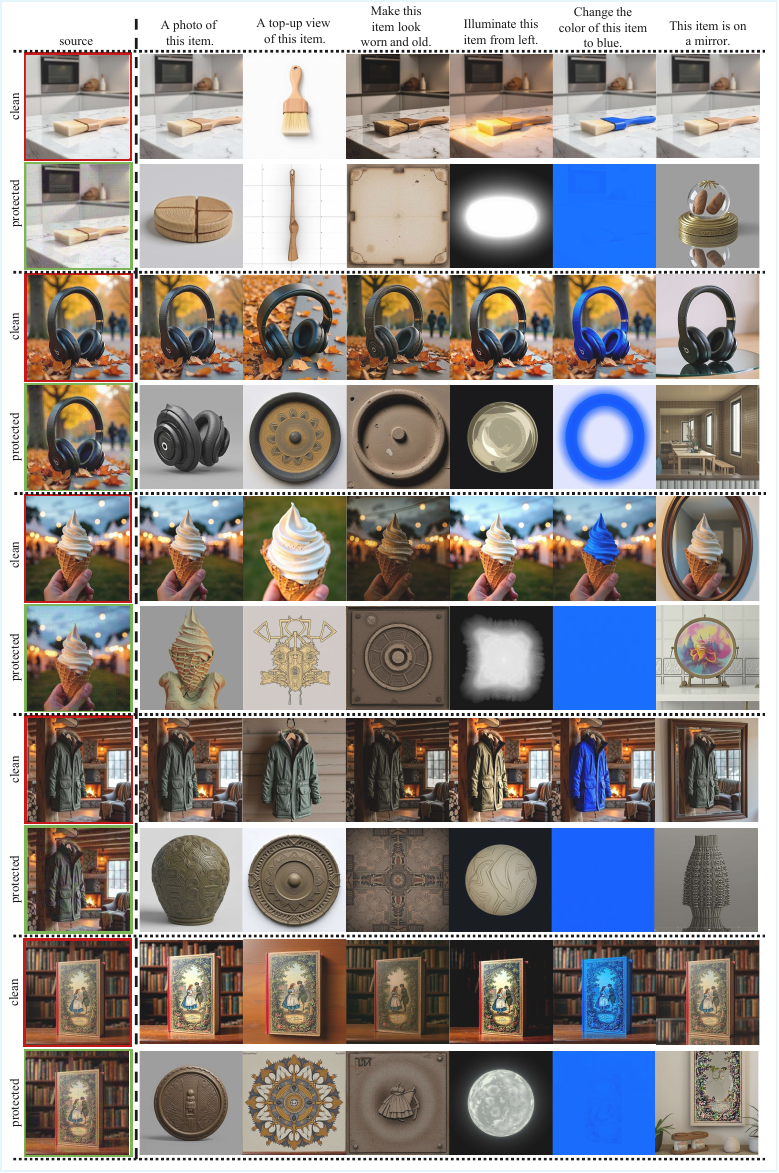}
    \caption{Defense on item images.}
    \label{Fig:item_results}
\end{figure*}

We provide more visualization results of human portraits defense (Fig.\ref{Fig:human_results}) and item images defense (Fig.\ref{Fig:item_results}).


